\documentclass[letterpaper, 10 pt, conference]{ieeeconf}  % Comment this line out if you need a4paper

\IEEEoverridecommandlockouts                              % This command is only needed if 
\usepackage{multicol}
\usepackage[bookmarks=true]{hyperref}
\usepackage{graphicx}
\usepackage{hyperref}
\usepackage{multirow}
\usepackage{threeparttable}
\usepackage{color,xcolor,soul}

\usepackage{amsmath}
\usepackage{amssymb}
\usepackage{mathtools}
\usepackage{amsthm}
\usepackage{adjustbox}
\usepackage{subcaption}
\usepackage{cite}

\usepackage[capitalize,noabbrev]{cleveref}
\DeclareMathOperator*{\minimize}{minimize}
\DeclareMathOperator{\tr}{tr}

\theoremstyle{plain}
\newtheorem{theorem}{Theorem}[section]
\newtheorem{example}[theorem]{Example}

\renewcommand{\baselinestretch}{0.987}

\begin{document}

% paper title
\title{A New Semidefinite Relaxation for Linear and Piecewise-Affine Optimal Control with Time Scaling}

% You will get a Paper-ID when submitting a pdf file to the conference system
\author{Lujie Yang, Tobia Marcucci, Pablo A. Parrilo, and Russ Tedrake\\
Department of Electrical Engineering and Computer Science \\
Massachusetts Institute of Technology
\thanks{Correspondence to: Lujie Yang \texttt{$<$lujie@mit.edu$>$}.}
\thanks{This work was supported by ONR N00014-22-1-2121.}
}
\maketitle
\begin{abstract}
We introduce a semidefinite relaxation for optimal control of linear systems with time scaling.
These problems are inherently nonconvex, since the system dynamics involves bilinear products between the discretization time step and the system state and controls.
The proposed relaxation is closely related to the standard second-order semidefinite relaxation for quadratic constraints, but we carefully select a subset of the possible bilinear terms and apply a change of variables to achieve empirically tight relaxations while keeping the computational load light.
We further extend our method to handle piecewise-affine (PWA) systems by formulating the PWA optimal-control problem as a shortest-path problem in a graph of convex sets (GCS).
In this GCS, different paths represent different mode sequences for the PWA system, and the convex sets model the relaxed dynamics within each mode.
By combining a tight convex relaxation of the GCS problem with our semidefinite relaxation with time scaling, we can solve PWA optimal-control problems through a single semidefinite program.
\end{abstract}
\IEEEpeerreviewmaketitle

\section{Introduction}
% Optimal control of linear and hybrid systems with time scaling is challenging due to the inherent nonconvexity of the problem. When the time scaling is fixed, optimizing the trajectory of a linear system can be effectively approached through convex programming. Alternatively, if the linear system's trajectory is predetermined, the time scaling problem can also be addressed within a convex optimization framework. However, the problem becomes considerably more complex when the trajectory and time scaling need to be optimized jointly. This bilinear interdependence between the trajectory parameters and the time scaling introduces significant challenges, as many approaches rely on nonconvex trajectory optimization that are prone to converging to local minima.
We consider the optimal-control problem (OCP) for discrete-time dynamical systems obtained by discretizing a continuous-time system with time step $h > 0$. 
Consider, for example, a system with continuous-time linear dynamics approximated with the Euler method: 
\begin{equation}\label{eq:lti_dynamics}
    x_{k+1} = x_k + h(Ax_k + Bu_k),
\end{equation}
where $x_k$ and $u_k$ are the state and control at discrete time $k$, and $A$ and $B$ are the matrices of the continuous-time system.
When the time scaling $h$ is fixed, optimizing the trajectory can be efficiently approached through convex optimization.
Similarly, if the system trajectory is predetermined, the time-scaling problem can also be addressed as a convex program.
However, if the trajectory and time scaling need to be optimized jointly, then the problem becomes nonconvex and considerably more complex to solve.
In fact, most common approaches rely on nonconvex trajectory optimization and are prone to converging to local minima.

Many existing works on trajectory optimization~\cite{betts1998survey,kelly2017introduction} generally require a predefined final time~\cite{garcia1989model,camacho2007model}, thereby limiting their ability to simultaneously optimize both the time scaling and trajectories. This restriction often overlooks low-cost trajectories that could emerge from varying terminal times, leading to less efficient or unsuitable solutions for time-critical applications~\cite{foehn2021time, dong2023time}. Additionally, the fixed-duration requirement is impractical for systems that can only reach the final state within a specific time frame. 
To remedy this inflexibility, some methods adopt a bilevel optimization approach~\cite{tang2019time,marcucci2024fast,kapania2016sequential,richter2016polynomial}, alternating between optimizing the time scaling on a fixed path and refining the path for a given time scaling. However, these algorithms are sensitive to the initial guess and easily get stuck in local optima. Direct-collocation techniques \cite{hargraves1987direct, kelly2017introduction} offer an alternative by parameterizing trajectories with splines and support incorporating time as a decision variable, but they rely on nonlinear programming (NLP) solvers that may also struggle with local optima or fail to identify feasible trajectories.
\begin{figure}
    \centering
\includegraphics[width=0.45\textwidth]{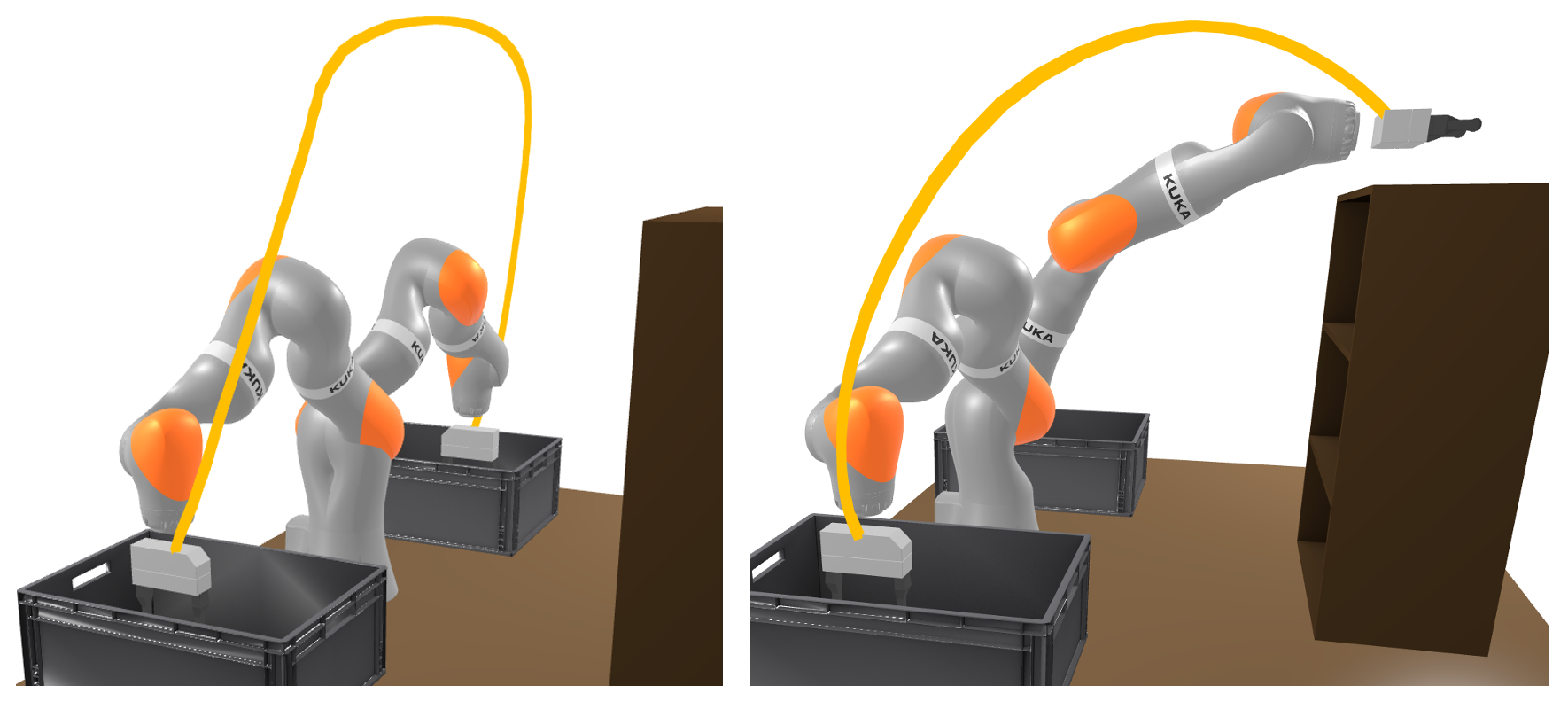}
        \caption{Minimum-time trajectories for a 7D double integrator moving among obstacles with robot arm geometry. The collision-free configuration space is approximately decomposed into convex polytopes. Our algorithm efficiently navigates these polytopes while respecting input limits. Videos of the trajectories can be found at~\href{https://lujieyang.github.io/ctfoc/comparison_LB_RB.html}{(left)} and~\href{https://lujieyang.github.io/ctfoc/comparison_RB_AS.html}{(right)}.\looseness=-1}
        \label{fig:iiwa_banner}
\end{figure}

Mixed-integer formulations for minimum-time control have been studied over the past decades~\cite{richards2006robust, carvallo1990milp, rothwangl2001numerical, richards2002aircraft}. By overcoming the localness of nonlinear programming, mixed-integer convex programming (MICP) has been applied to motion planning for autonomous vehicles~\cite{schouwenaars2001mixed,deits2015efficient,landry2016aggressive}, locomotion~\cite{deits2014footstep,kuindersma2016optimization,aceituno2017simultaneous}, and manipulation~\cite{blackmore2006optimal, ding2011mixed}. MICP approaches often handle the originally nonconvex constraints with piecewise convex approximations, and can be solved to global optimum given enough time budget. Additionally, MICP methods are naturally capable of generating optimal trajectories or feedback controllers for piecewise-affine (PWA) dynamical systems with contacts~\cite{marcucci2017approximate,hogan2016feedback,marcucci2019mixed}. These strategies typically adopt integer variables to assign system modes, enabling the simultaneous planning of discrete mode sequences and continuous motions. 
% For instance, Aceituno et al. approximate the robot torque limits and friction cone constraints with polytopic constraints, and optimize contacts and trajectories for legged robots~\cite{aceituno2020global} and robotic manipulators~\cite{aceituno2017simultaneous}.
% In~\cite{marcucci2019mixed}, a thorough examination of various MICP formulations explores the balance between computational demand and representation power, with applications demonstrated in hybrid model predictive control~\cite{borrelli2017predictive,camacho2010model}.
Nevertheless, these methods typically require integer variables to decide the system mode at every time step for a predefined horizon, therefore imposing considerable computational challenges. 

Recently, the work~\cite{marcucci2024shortest} has proposed an efficient MICP formulation of hybrid OCPs as finding the shortest paths in a Graph of Convex Sets (GCS) (see also~\cite[Ch.~10]{marcucci2024graphs}). The GCS approach enjoys a tight convex relaxation of the nonconvex control problem. By addressing the planning problem through a single convex program, GCS empowers a range of effective planners~\cite{marcucci2023motion,cohn2023non,graesdal2024towards} capable of handling high-dimensional manipulation and contact problems. 

In this paper, we leverage semidefinite relaxation (SDR)~\cite{parrilo2000structured,parrilo2003semidefinite,lasserre2001global,luo2010semidefinite} to convexify the nonconvex dynamical constraints and incorporate them in the GCS framework. We additionally allow the time step to be decision variables to jointly optimize trajectories and time scaling for PWA systems. Recent advances in SDRs have enabled applications to collision-free motion planning~\cite{el2021piecewise,teng2023convex,zhang2021semi} and optimal control~\cite{lasserre2008nonlinear,jiang2015global,yang2023approximate,kang2024fast}. However, these methods have mostly been limited to path planning in low-dimensional spaces or smooth dynamics without mode switches. Moreover, previous studies~\cite{teng2023convex,kang2024fast} found that a computationally expensive complete higher-order SDR is necessary to tighten the relaxation with dynamical constraints. We show that by leveraging the specific structure of the system dynamics, our new formulation achieves empirical tightness with a smaller, more efficient relaxation, removing redundant terms and significantly improving computational efficiency.\looseness=-1

SDRs also play an important role in other robotics problems, including inverse kinematics~\cite{trutman2022globally,maric2020inverse} and certifiable perception algorithms~\cite{kahl2007globally,yang2022certifiably}. Similar to our work, \cite{graesdal2024towards} uses SDR with GCS, but to relax the nonconvex quasi-static dynamics and solves a planar pushing task for fixed final times. In our work, we exploit the structure of the OCP with time scaling, applying and adapting the SDR technique to generate globally optimal trajectories for linear systems. We then further explore such relaxation's potential in trajectory optimization for PWA dynamical systems using GCS.

Our key contributions are summarized as follows:

\begin{enumerate}
\item We propose an SDR for nonconvex OCPs for linear systems with time scaling.
This SDR exploits the problem structure to be both empirically tight and lightweight.
\item We extend the SDR to handle PWA systems by transcribing the mode optimization problem as a shortest-path problem in GCS. Our method simultaneously optimizes the mode sequence and system trajectory, within a unified convex-optimization framework.
\end{enumerate}
We validate our approach on an inverted pendulum with contact and a 7D double integrator approximation of the dynamics of a robot arm (see Fig.~\ref{fig:iiwa_banner}). By decomposing the nonconvex OCP into convex components, we are able to jointly optimize trajectories and time scaling for PWA systems that are otherwise challenging for traditional NLP or MICP solvers.

\section{Preliminaries on Semidefinite Relaxation}
In this section, we provide the necessary background on quadratically constrained quadratic programs (QCQP) and their SDR. The interested readers can refer to~\cite{luo2010semidefinite}, as well as~\cite{parrilo2000structured,parrilo2003semidefinite,lasserre2001global}, for more details.

Given the variable $z \in \mathbb{R}^n$, we consider the following QCQP (in homogeneous form in terms of $x = (1,z)$):
\begin{subequations} \label{eq:qcqp}
    \begin{align}
        \minimize  \quad x^\top A_0 x \\
        \text{subject to} \quad x^\top A_i x &\ge 0, \quad \forall i = 1, \ldots, l \\
         Bx&\ge 0\\
         x_1&=1,
    \end{align}
\end{subequations}
where $A_0, \ldots, A_l \in \mathbb{R}^{(n+1)\times(n+1)}$ are symmetric matrices, $B \in \mathbb{R}^{m\times(n+1)}$, and $x_1$ denotes the first entry of $x$.
To handle the nonconvex QCQP \eqref{eq:qcqp}, the standard SDR introduces a positive semidefinite (PSD) matrix variable $X$, which serves as a relaxation of $xx^\top$. By dropping the nonconvex rank-1 constraint $X=xx^\top$, we arrive to the semidefinite program
\begin{subequations}\label{eq:standard_sdr}
    \begin{align}
        \minimize  \quad \tr(A_0 X) \\
        \text{subject to} \quad \tr(A_iX) &\ge 0, \quad \forall i = 1, \ldots, l \\
         BXe_1&\ge 0\\
         BXB^\top &\ge 0 \\
         e_1^\top X e_1 &= 1 \\
         X &\succeq 0,
    \end{align}
\end{subequations}
where $e_1 = (1,0, \ldots, 0) \in \mathbb R^{n+1}$.  The optimal value of the SDR~\eqref{eq:standard_sdr} provides a lower bound for the original QCQP \eqref{eq:qcqp}. If the relaxation were exact (i.e., if $X$ is rank-1), the optimal solution of \eqref{eq:qcqp} could be directly obtained by factorizing  $X$ as $X = xx^\top$. However, in general, the SDR may yield a higher-rank solution, and additional steps may be needed to recover a feasible solution for the original nonconvex QCQP.
\section{Problem Statement}
\label{section: formulation}
We consider a discrete-time dynamical system obtained by discretizing a continuous-time system with time step $h > 0$.
The system state $x_k \in \mathbb R^{n_x}$ and control input $u_k \in \mathbb R^{n_u}$ at each discrete time $k$ are constrained to lie in the polytopes $\mathcal X \subseteq \mathbb R^{n_x} $ and $\mathcal U \subseteq \mathbb R^{n_u}$.
We aim to design optimal state trajectories $\{x_k\}_{k=0}^K$ and control inputs $\{u_k\}_{k=0}^{K-1}$, alongside the time step $h$.
Note that the time step $h$ directly parameterizes the time scaling of the entire trajectory.
The objective function balances the total time $Kh$ and system energy with the weight $\eta \ge 0$.
The initial state is given by $x_\text{start}$, and the final state is required to be $x_\text{goal}$.
Overall, the OCP is \looseness=-1
\begin{subequations} \label{eq:general_opt}
\begin{align}
    \minimize \quad & \eta Kh + h \sum_{k=0}^{K-1}  (x_k^{\top}Qx_k + u_k^{\top}Ru_k )\label{eq:opt_ctrl_obj}\\ 
    \text{subject to} \quad & x_{k+1} = f(x_k, u_k, h), \quad \forall k = 0, \ldots, K-1 \\
    & x_k \in \mathcal{X}, \ u_k \in \mathcal{U}, \quad \forall k = 0, \ldots, K-1 \\
    &x_0 = x_{\text{start}}, \; x_{K} = x_{\text{goal}},
\end{align}
\end{subequations}
where the matrices $Q, R$ are PSD, and the function $f$ represents the discrete-time system dynamics. 

% \textbf{Linear Time Invariant System}
% For instance, in case of a linear time-invariant (LTI) system with continuous-time matrices $A$ and $B$, the dynamics derived from the forward Euler method is
% \begin{equation}\label{eq:lti_dynamics}
%     f(x_k, u_k, h) = x_k + h (Ax_k+ B u_k).
% \end{equation}

\section{Convex Relaxation with Time Scaling}
Even for LTI systems, the OCP~\eqref{eq:general_opt} is nonconvex due to the biconvex interaction between trajectory parameters $\{x_k\}_{k=0}^K$, $\{u_k\}_{k=0}^{K-1}$, and the time step $h$ (see the objective~\eqref{eq:opt_ctrl_obj} and the dynamics~\eqref{eq:lti_dynamics}). To tackle this challenge, we propose a convex relaxation of the nonconvex program that leverages the problem structure.
We group the bilinear terms to construct an expressive SDR that aligns with the OCP structure.
Our approach is first demonstrated for LTI systems, which represent the simplest framework for understanding the logic behind our method.
Then, in Section~\ref{sec:mode_fixed}, we expand this strategy to handle PWA systems with a fixed mode schedule. Finally, in Section~\ref{sec:mode_scheduling}, we discuss how to also optimize the mode sequence for PWA systems.

% We start with a discrete-time linear time-invariant (LTI) system derived from the forward Euler method
% \begin{subequations} \label{eq:lti_prog}
% \begin{align}
%     \minimize \quad & Kh + h \sum_{k=0}^{K-1}  x_k^{\top}Qx_k + u_k^{\top}Ru_k \label{eq:opt_ctrl_obj}\\ 
%     \text{subject to  } & x_{k+1}- \quad \forall t = 0, \ldots, K-1\\
%     & x_k \in \mathcal{X}, u_k \in \mathcal{U} \\
%     &x_0 = x_{\text{start}}, \; x_{K} = x_{\text{goal}},
% \end{align}
% \end{subequations}
% \tmcomment{Use minimize and not min. Also do not write the variables under min, they clutter the equations too much. Typically, the first sentence after the optimization problem says what are the variables.}
% where $A$ and $B$ are the system matrices in continuous time. 
\subsection{Time-Flexible Relaxation}
% \tmcomment{From Pablo: did we try writing $h = w^2$, and doing a change of variables $x’ = wx$ and $u’ = wu$?}
% \tmcomment{From Pablo: Regardless of this, my sense is that it’d be nice to try to present the gist of the problem/solution a bit more clearly (e.g., along the lines of eqs. (12), but perhaps that’s in the works already.  This may help us understand if “we’re leaving money on the table}

The ordinary SDR for nonconvex QCQPs~\cite{luo2010semidefinite} is inadequate for addressing~\eqref{eq:general_opt} since the stage cost \eqref{eq:opt_ctrl_obj} is not quadratic but rather cubic in the decision variables.
To handle the polynomial terms $hx_k^{\top}x_k, hu_k^{\top}u_k, hx_k, hu_k$ and obtain a tight convex relaxation of the nonconvex program, we define $r = (x_0, \ldots, x_K, u_0, \ldots, u_{K-1})$.
We then introduce the following vector $y$ and its corresponding PSD matrix $Y$:
\begin{equation}\label{eq:change_of_variable}
y = \begin{bmatrix}
1 \\
h \\
r \\
hr
\end{bmatrix}, \quad Y = \frac{1}{h} yy^\top = \begin{bmatrix}
\frac{1}{h} & 1& \frac{r^\top}{h} & r^\top \\
1 & h & r^\top & hr^\top  \\
\frac{r}{h} & r& \frac{rr^\top}{h} & rr^\top  \\
r & hr & rr^\top & hrr^\top  
\end{bmatrix}.
\end{equation}
The construction of $Y$ exploits the fact that the time step $h$ is positive, and is tailored to the bilinear structure of the OCP.
Notice that $y$ includes the second-order term $hr$ but excludes $h^2$ and $r\otimes r$, which would be part of an ordinary second-order SDR~\cite{parrilo2003semidefinite,lasserre2001global}.
The multiplication by $1/h$ in the matrix $Y$ transforms the terms $h^2rr^\top$ and $h^2$, which do not appear in the OCP \eqref{eq:general_opt}, into $hrr^\top$ and $h$, which are necessary to make the objective~\eqref{eq:opt_ctrl_obj} linear in $Y$.

We report the detailed time-flexible relaxation (TFR) program in the Appendix, see~\eqref{eq:lti_tsr}.
There, additional constraints obtained by multiplying linear constraints of the original QCQP are added.
While these constraints are redundant for the original nonconvex formulation, they play a crucial role in tightening the SDR.

\subsection{Sparse Time-Flexible Relaxation}
Although the TFR provides empirically tight solutions to all the control problems considered in this paper, it can be computationally expensive. Specifically, the number of variables in $y$ is doubled compared to $r$, which consequently squares the number of variables in $Y$. Therefore this dense SDR can present significant scalability challenges for more complex systems. To address these challenges, we leverage the sparsity in the OCP and alleviate the computational demands. In particular, we exploit the fact that only the specific group of variables $(x_k, x_{k+1}, u_k)$ are coupled in the costs and constraints.
This allows us to build an SDR using the variables
\begin{equation}
r_k = \begin{bmatrix}x_k \\ x_{k+1} \\u_k \end{bmatrix}, \quad 
y_k = \begin{bmatrix} 1 \\ h \\r_k \\hr_k\end{bmatrix}, \quad 
Y_k = \frac{1}{h} y_ky_k^\top.
\end{equation}
To ensure state variables are consistent across time steps, we impose the equality constraints
\begin{equation}\label{eq:state_consistency}
    r_k [n_x : 2 n_x] = r_{k+1}[:n_x],
\end{equation}
where the notation follows the Python indexing rules. Additional constraints that further tighten the relaxation are detailed in the Appendix, see~\eqref{eq:state_consistency_additional}. 
\begin{figure}
\centering
\includegraphics[width=0.48\textwidth]{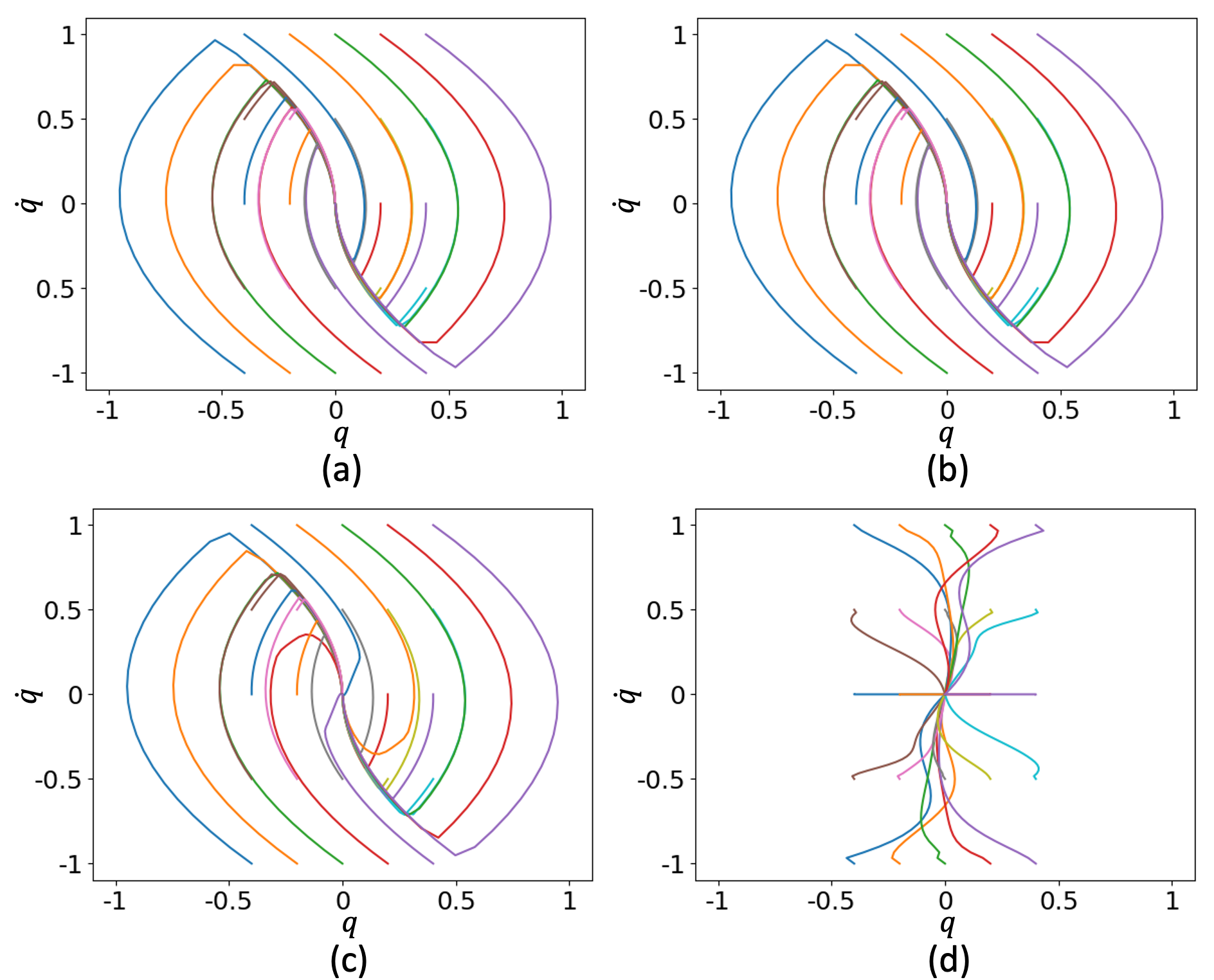}
    \caption{Comparison of different SDRs for the minimum-time double integrator. (a) Ground truth state-space trajectories. (b) Trajectories from dense TFR. (c) Trajectories from sparse TFR. (d) Trajectories from standard SDR.}
	\label{fig:sdr_comparison}
\vspace{-0.2cm}
\end{figure}
\begin{example}[Minimum-time double integrator]
Fig.~\ref{fig:sdr_comparison} compares the tightness of different SDRs on a minimum-time double integrator problem, whose optimal state trajectories are shown in Fig. ~\ref{fig:sdr_comparison}(a). In this example, $K$ is set to 30, and the minimum-time objective excludes the quadratic state and control penalties ($Q=R=0$), making it amenable to the standard first-order SDR. The standard dense first-order SDR uses a 94-dimensional PSD matrix and results in a loose approximation of the closed-loop trajectories (Fig.~\ref{fig:sdr_comparison}(d)). Although a dense second-order SDR could potentially provide tighter relaxations, it requires a 4465-dimensional PSD matrix for this problem and runs out of memory due to the large $K$. In contrast, our dense TFR only requires a 186-dimensional PSD matrix and precisely recovers the optimal controller (Fig.~\ref{fig:sdr_comparison}(b)). Moreover, our sparse TFR uses 29 PSD matrices of dimension 12 and produces trajectories that closely match the optimal policy (Fig.~\ref{fig:sdr_comparison}(c)), while being computationally more efficient than the dense formulation.
\end{example}

\section{Piecewise-Affine Systems with a Fixed Mode Sequence}
\label{sec:mode_fixed}

We now extend our TFR to handle optimal control of discrete-time PWA systems of the form
\begin{multline}
\label{eq:pwa_dynamics}
x_{k+1} = x_k + h (A_i x_k + B_i u_k + c_i) \\
\text{if} \ (x_k, u_k)\in \mathcal{X}_i \times \mathcal{U}_i,
\end{multline}
with the system mode $i\in\{1, \ldots, s\}$ and polytopic domains $\mathcal{X}_i \in \mathbb{R}^{n_x}, \mathcal{U}_i \in \mathbb{R}^{n_u}$.
Here, we assume that we are given the number of mode switches $N$ and the fixed mode sequence $i_1, \ldots, i_N$ traversed by the system (the optimization of the mode sequence will be discussed in Section~\ref{sec:mode_scheduling}). We divide the trajectory into $N$ segments.
In each segment, the system evolves for $K_n$ time steps under a specific mode. The objective is to find the optimal trajectory $\{x_k^n\}_{k=0}^{K_n}, \{u_k^n\}_{k=0}^{K_n-1}$ and time duration $h_n$ in each mode $n\in \{1,\ldots,N\}$.
This results in the optimization problem 
\begin{subequations} \label{eq:pwa_prog}
\begin{align}
\minimize \quad  &
% \eta \sum_{n=1}^N K_n h_n + \sum_{n=1}^N  h_n \sum_{k=0}^{K_n-1} ({x_k^n}^{\top}Qx_k^n + {u_k^n}^{\top}Ru_k^n) \label{eq:pwa_cost}\\
\sum_{n=1}^N h_n \left( \eta K_n + \sum_{k=0}^{K_n-1} ({x_k^n}^{\top}Qx_k^n + {u_k^n}^{\top}Ru_k^n) \right) \label{eq:pwa_cost}\\
\text{subject to} \quad & \forall  n = 1, \ldots, N, \ \forall k = 0, \ldots, K_n-1: \nonumber \\
& x_{k+1}^n = x_k^n + h_n(A_{i_n}x_k^n+ B_{i_n} u_k^n + c_{i_n})  \label{eq:pwa_dynamics}\\
& x_k^n \in \mathcal{X}_{i_n}, \ u_k^n \in \mathcal{U}_{i_n} \label{eq:pwa_state_input_constr}\\
& x_0^n = x_{K_{n-1}}^{n-1} \label{eq:pwa_state_continuity}\\
&x_0^1 = x_{\text{start}}, \; x_{K_N}^N = x_{\text{goal}}. \label{eq:pwa_boundary_cond}
% \vspace{-0.1cm}
\end{align}
\end{subequations}
Constraint~\eqref{eq:pwa_state_continuity} ensures that the state trajectory is continuous during the mode transition. Instead of applying the TFR to the entire program~\eqref{eq:pwa_prog}, we modularize the computation and take the TFR of~\eqref{eq:pwa_cost}--\eqref{eq:pwa_state_input_constr} in each mode independently.
This TFR is denoted as $\mathrm{S}_n$, and has variables $Y^n = y^n {y^n}^\top / h_n$ where $r^n = (x_0^n, \ldots, x_K^n, u_0^n, \ldots, u_{K-1}^n)$ and $y^n = (1, h_n, r^n, h_nr^n)$.
The boundary conditions \eqref{eq:pwa_boundary_cond} are imposed in the initial and terminal mode, respectively, as in~\eqref{eq:12f}--\eqref{eq:12k}. 

We highlight that this modular relaxation significantly reduces the number of binary variables that parameterize a mode switch, from being proportional to the total number of time steps (as in traditional MICP formulations ~\cite{marcucci2019mixed}) to being proportional to the number of mode switches.
This significantly reduces the computational cost of the MICP transcription.\looseness=-1

\subsection{Continuity Constraints with Coupling Matrix}
To effectively enforce the state continuity constraints \eqref{eq:pwa_state_continuity}, we introduce a vector $\bar y$ that couples the time steps and equivalent states in consecutive modes.
For all $n = 1, \ldots, N-1$, we define \looseness=-1
\begin{equation}\label{eq:coupling_mtrx_def}
\begin{aligned}
\bar x &= x_{K_n}^n = x_0^{n+1}, \
\bar Y = \bar y \bar y^\top \\
 \bar y &= \left(
            1 , h_{n} , h_{n+1} , \bar{x} , h_{n} \bar{x} ,h_{n+1} \bar{x}, \frac{\bar{x}}{h_{n}} , \frac{\bar{x}}{h_{n+1}}
       \right).
    \end{aligned}
\end{equation}
The coupling PSD matrix $\bar{Y}$ includes terms such as:
\begin{align*}
& \bar{x}\bar{x}^\top, \ h_n \bar{x}, \ h_{n+1} \bar{x}, \ \bar{x} / h_n, \ \bar{x} / h_{n+1}, \\
& h_n \bar{x}\bar{x}^\top, \ h_{n+1}\bar{x}\bar{x}^\top, \ \bar{x}\bar{x}^\top /h_n, \ \bar{x}\bar{x}^\top / h_{n+1},
\end{align*}
% {\small
% \begin{align*}
% \bar{x}\bar{x}^\top, &\ h_n \bar{x}, \ h_{n+1} \bar{x}, \ \frac{\bar{x}}{h_n}, \ \frac{\bar{x}}{h_{n+1}}, \
% h_n \bar{x}\bar{x}^\top, \ h_{n+1}\bar{x}\bar{x}^\top, \ \frac{\bar{x}\bar{x}^\top}{h_n}, \ \frac{\bar{x}\bar{x}^\top}{h_{n+1}},
% \end{align*}}%
which are designed to precisely match the corresponding entries in $Y^n$ and $Y^{n+1}$ of consecutive modes.
This ensures that both the state variables and the time-dependent elements of the system's dynamics are accurately aligned across the mode transitions, effectively tightening the SDR. 
The coupling matrix $\bar{Y}$ allows us to enforce crucial additional constraints on the PSD matrices $Y^n$ paired with consecutive modes, especially those involving $h_n$, which are critical for maintaining system continuity and optimizing the overall trajectory.\looseness=-1

\begin{figure}
\centering
\includegraphics[width=0.48\textwidth]{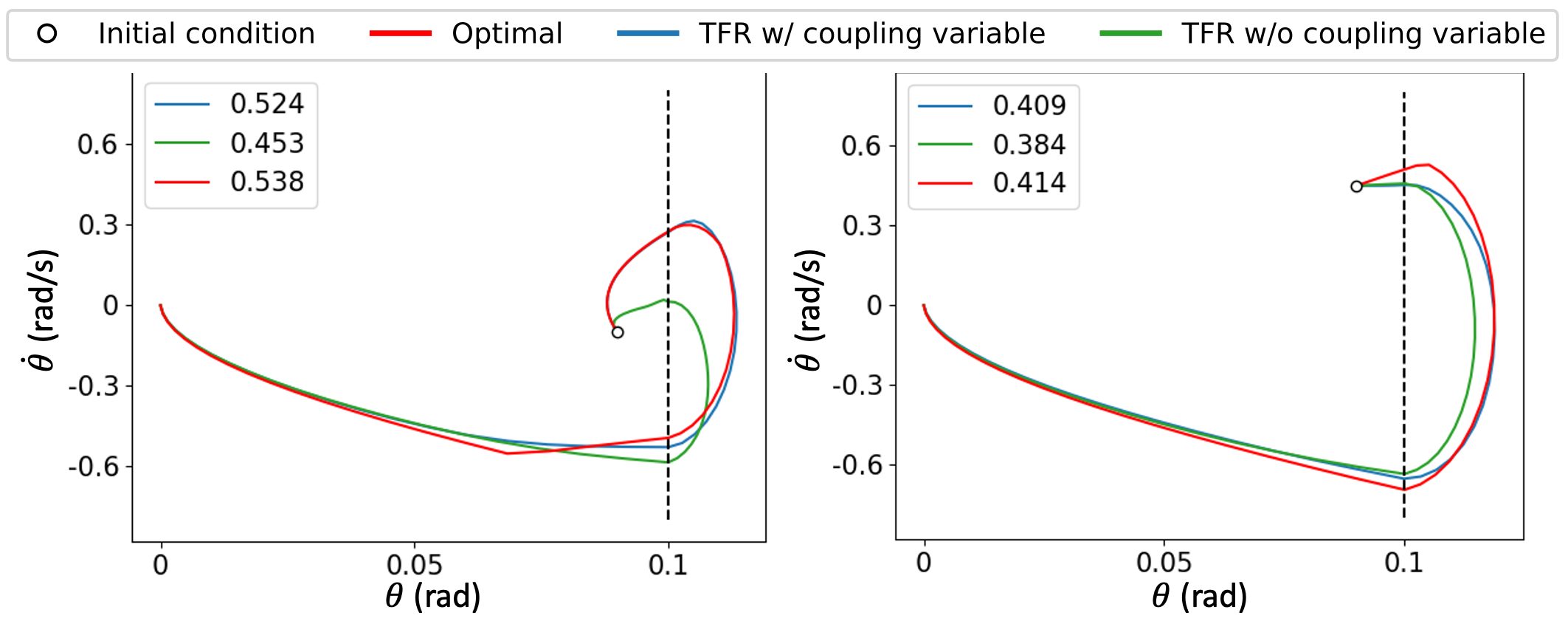}
    \caption{Effect of the coupling variable on balancing the inverted pendulum with the wall in minimum time for different initial conditions. The vertical dashed line indicates the angular limit beyond which the pendulum collides with the wall. The legend indicates the time duration for each trajectory.\looseness=-1}
\label{fig:coupling_variable_comparison}
\vspace{-0.3cm}
\end{figure}

\begin{example}[Minimum-time Inverted Pendulum with Wall] Fig.~\ref{fig:coupling_variable_comparison} illustrates the effectiveness of the coupling matrix $\bar Y$ in adding useful continuity constraints. The task is to balance a linearized inverted pendulum around the upright equilibrium between elastic walls in minimum time. The system parameters are adopted from~\cite{marcucci2017approximate}, with the state defined as $x = (\theta, \dot \theta)$. For different initial conditions, the TFR trajectory that incorporates the coupling variable consistently provides a tighter lower bound to the optimal solution. 
\label{example:min_time_pendulum_w_wall}
\end{example}

\section{Mode Optimization for Piecewise-Affine Systems}
\label{sec:mode_scheduling}

In the previous section, we discussed how to generate optimal trajectories for PWA systems with a predefined mode sequence.
Here, we show how to optimize the system trajectory and the mode sequence jointly, while respecting the dynamics in each mode.
We think of the different mode sequences as different paths in a graph of modes, and we formulate the OCP as a shortest-path problem in GCS~\cite{marcucci2024shortest}.
The convex sets and functions in this GCS will be constructed leveraging our TFR.

\subsection{Shortest Paths in Graphs of Convex Sets}
\label{subsec:gcs}

A GCS is a directed graph $G= (\mathcal{V}, \mathcal{E})$ with vertex set $\mathcal{V}$ and edge set $\mathcal{E}$.
Each vertex $v\in \mathcal{V}$ of a GCS is associated with a convex compact set $\mathcal{X}_v$ and a continuous variable $x_v$ inside it.
Instead of being a constant as in an ordinary weighted graph, the length of an edge $e = (u, v) \in \mathcal E$ in a GCS is a nonnegative convex function $l_e(x_u, x_v)$  of the continuous variables 
paired with the edge endpoints.
Each edge is also paired with a convex constraint of the form $(x_u, x_v) \in \mathcal{X}_e$.

The shortest-path problem in a GCS is formulated as follows: \looseness=-1
\begin{subequations}
\begin{alignat}{2}
\minimize \quad & \sum_{e= (u, v) \in\mathcal{E}_p} l_e(x_u, x_v) \\
\text{subject to} \quad & p \in \mathcal{P} \\
& x_v \in \mathcal{X}_v, &&\quad \forall v \in\mathcal{V}\\
& (x_u, x_v) \in \mathcal{X}_e, \quad && \forall e= (u, v) \in\mathcal{E}_p,
\end{alignat}
\end{subequations}
where $p$ is a path between two pre-specified vertices in $G$, $\mathcal{P}$ is the family of such paths, and $\mathcal{E}_p$ denotes the set of edges traversed by $p$.
The decision variables are both the discrete path $p$ and the continuous variables $x_v$ for all $v \in \mathcal V$.
The shortest-path problem in GCS is NP-hard~\cite[Theorem 1]{marcucci2024shortest}.
However, the techniques proposed in~\cite{marcucci2024shortest} allow us to derive an empirically tight convex relaxation of this problem.

We will model our PWA OCP as a shortest-path problem in GCS, where different paths represent different mode sequences.
Then we will use our TFR, concurrently with the techniques from~\cite{marcucci2024shortest}, to derive a unified convex relaxation of the PWA control problem.\looseness=-1

\subsection{Spectrahedra as Convex Sets $\mathcal{X}_v$}\label{subsec: spectrahedra_convex_set}
We associate each mode $i$ with a vertex $v \in \mathcal{V}$ in GCS and construct a spectrahedron (see, e.g.,~\cite[Sec.~2.1.2]{blekherman2012semidefinite}) as the feasible set of each semidefinite program $\mathrm{S}_i$. These spectrahedra encode the relaxed dynamical constraints, and serve as convex sets $\mathcal{X}_v$ in the GCS.
Similarly, their corresponding PSD matrices $Y^i$ serve as the continuous variables, $x_v$.

\subsection{Optimal Control of PWA Systems with Time Scaling}
% \begin{figure}
% \centering
% \includegraphics[width=0.4\textwidth]{figures/gcs_tfr.png}
%     \caption{Proposed workflow for optimal control of PWA systems with time scaling. Each TFR parameterizing a trajectory in each mode becomes a spectrahedron in the GCS.}
% 	\label{fig:tfto_pipeline}
%     % \vspace{-0.2cm}
% \end{figure}
We now outline the workflow of convex optimal control with time scaling for PWA systems. For each mode $i$, we construct the spectrahedron from $\mathrm{S}_i$ as the convex set and solve GCS to determine the mode sequence $p=(i_1, \ldots, i_N)$ with an approximate optimal trajectory $(r_{i_1}, \ldots, r_{i_N})$ satisfying relaxed dynamical constraints. We then refine the trajectory for the mode sequence $p$ using nonlinear programming solvers for the original bilinear program \eqref{eq:pwa_prog} (not the SDR) to obtain an optimal and dynamically-consistent trajectory from the good initial guess given by the combination of GCS and our TFR (henceforth GCS+TFR).

\section{Additional Experimental Results}
\label{section:results}
We demonstrate the effectiveness of our approach on an inverted pendulum with contact and a 7DoF robot arm. By convexifying the system dynamics and the mode scheduling problem, while enabling flexible time duration, we are able to generate optimal trajectories that are challenging for traditional NLP or MICP methods.

\subsection{Inverted Pendulum with Elastic Wall}
We apply our method to balance an inverted pendulum around the upright equilibrium between elastic walls as in Example \ref{example:min_time_pendulum_w_wall}.
We compare TFR with the fixed-time-step mixed-integer convex formulation for PWA systems proposed in~\cite{moehle2015perspective,marcucci2019mixed}. The MICP approach uses a time step of $h=0.005$ and $K = 160$ for a comparable discretization resolution, and can only find final times that are multiples of the fixed time step. In contrast, our method uses $K=20$ per mode and allows for finer, floating-point time steps. 
\begin{figure}
\centering
\includegraphics[width=0.49\textwidth]{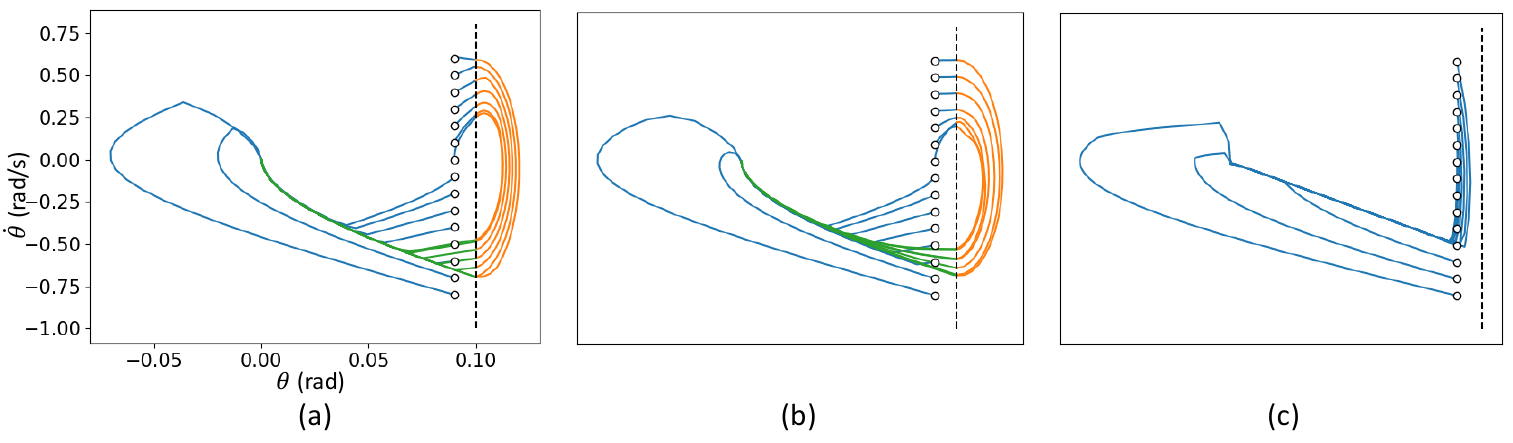}
    \caption{Time-optimal trajectories for the inverted pendulum with wall. (a) GCS+TFR+NLP solutions. The vertical dashed line represents the angle at which the pendulum makes contact with the wall. (b) GCS+TFR solutions.  (c) Convex relaxation of MICP formulation with fixed $h$. The different segment colors indicate mode transitions.}
    \label{fig:pendulum_with_wall}
    \vspace{-0.2cm}
\end{figure}
\begin{figure}
\centering
\includegraphics[width=0.48\textwidth]{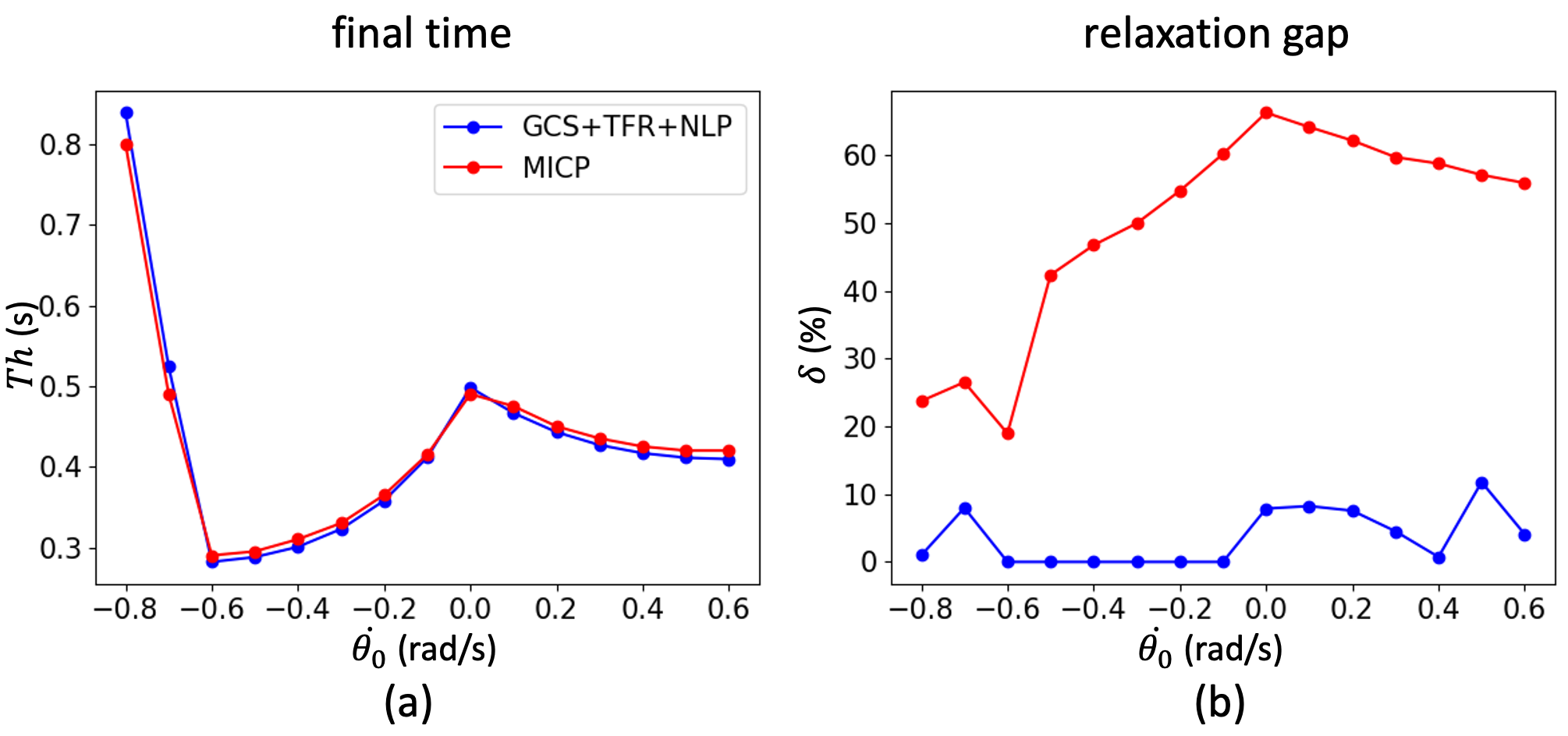}
    \caption{Performance comparison for balancing the inverted pendulum with contact in minimum time. The pendulum starts from the initial conditions $(0.09, \dot \theta_0)$. (a) Final time. (b) Relaxation gap. The proposed GCS+TFR formulation is much tighter than the relaxation of the standard MICP. }
\label{fig:pendulum_statistics}
\vspace{-0.2cm}
\end{figure}
Fig.~\ref{fig:pendulum_with_wall} shows the minimum-time trajectories for different initial conditions.
The GCS+TFR solution in Fig.~\ref{fig:pendulum_with_wall}(b) closely approximates the globally optimal solution refined by the NLP solver in Fig.~\ref{fig:pendulum_with_wall}(a).
In contrast, the convex relaxation of MICP in Fig.~\ref{fig:pendulum_with_wall}(c) is very loose, resulting in trajectories that consistently miss contacts with the wall. 
Fig.~\ref{fig:pendulum_statistics}(a) shows the minimum time required to balance the inverted pendulum using both methods. We compare the tightness of GCS+TFR with the convex relaxation of MICP in Fig.~\ref{fig:pendulum_statistics}(b). The relaxation gap for our method is defined as $\delta = (C_{\text{NLP}}-C_{\text{GCS}}) /C_{\text{NLP}}$, where $C_{\text{NLP}}$ and $C_{\text{GCS}}$ are the costs of GCS+TFR+NLP and GCS+TFR respectively. For MICP, $\delta = (C_{\text{opt}}-C_{\text{relax}}) /C_{\text{opt}}$, where $C_{\text{opt}}$ is the optimal cost and $C_{\text{relax}}$ is the cost of the relaxed program.

\begin{figure}
\centering
\includegraphics[width=0.49\textwidth]{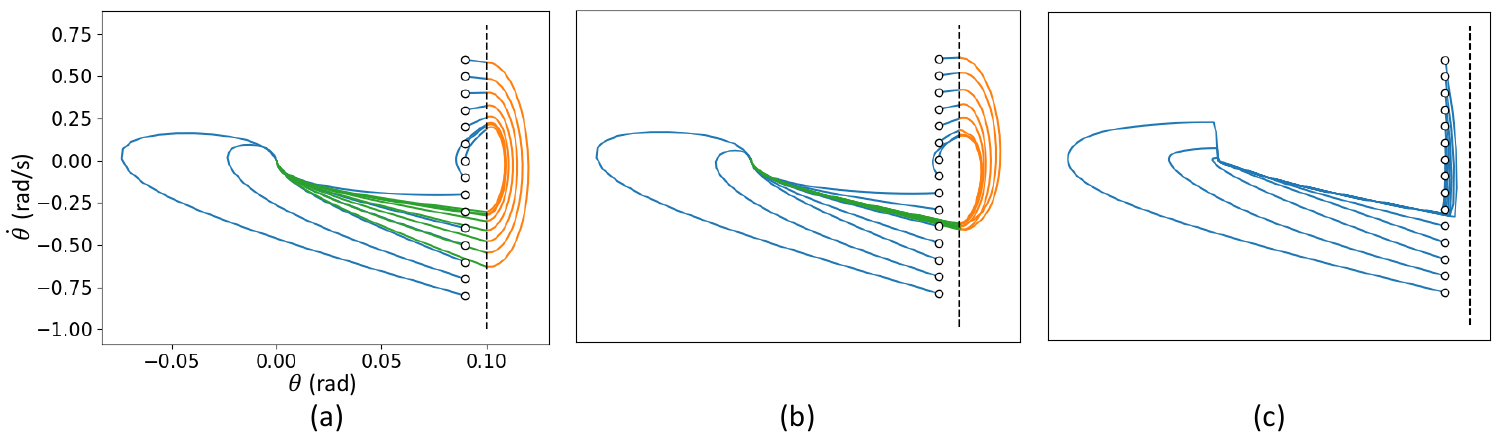}
    \caption{Trajectories for balancing the inverted pendulum with the time and control-effort cost. (a) GCS+TFR+NLP solutions. (b) GCS+TFR solutions. (c) Convex relaxation of MICP.
    }
    \label{fig:pendulum_1_1}
\end{figure}

\begin{figure}
\centering
\includegraphics[width=0.48\textwidth]{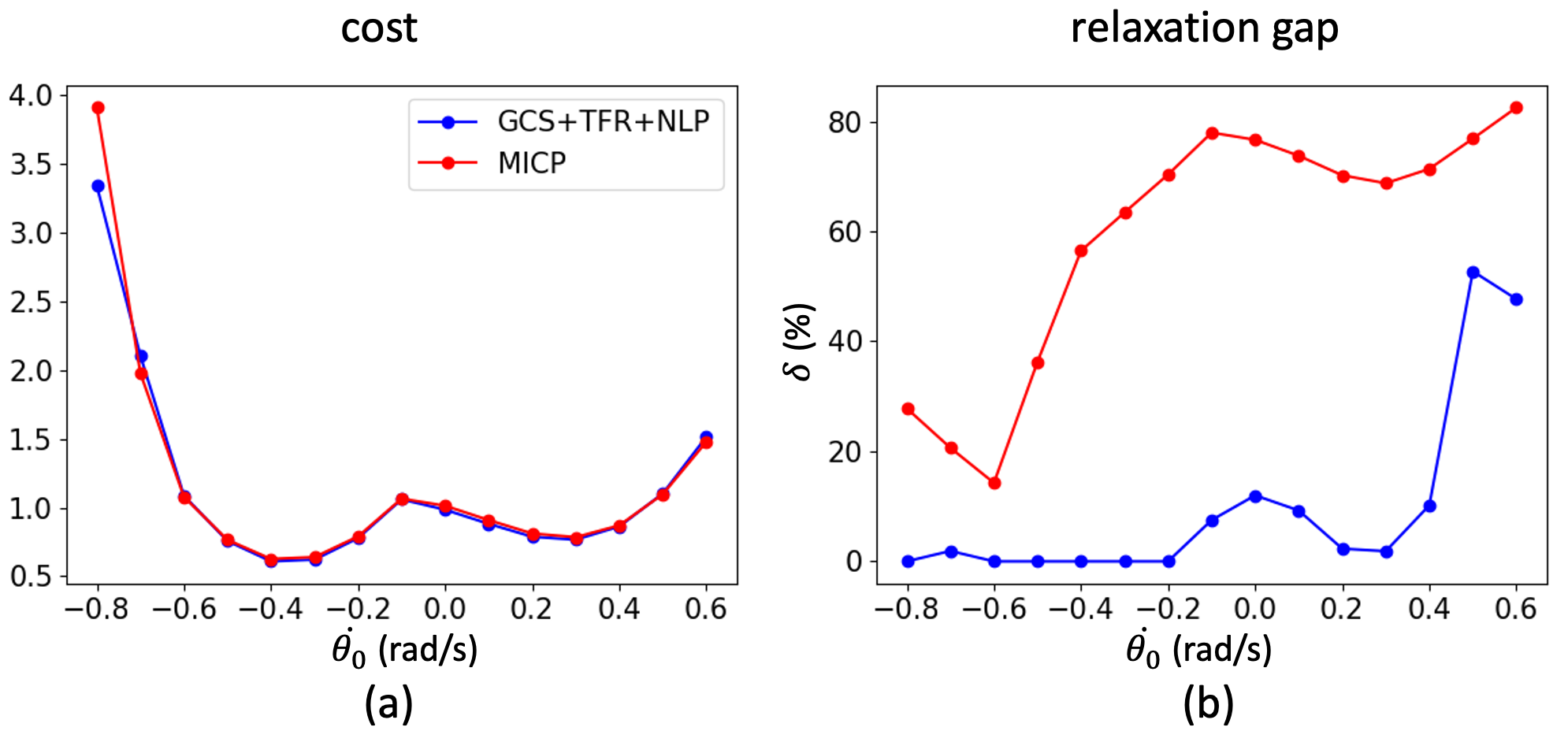}
    \caption{Performance comparison for balancing the inverted pendulum with a time and a control effort penalty.}
% \vspace{-0.2cm}
\label{fig:pendulum_statistics_1_1}
\end{figure}

Our approach is not limited to minimum-time problems.
To show this, in Fig.~\ref{fig:pendulum_1_1} we report the trajectories generated to stabilize the pendulum with both  time and control-effort cost ($\eta = R= 1$ and $Q = 0$).
Fig.~\ref{fig:pendulum_statistics_1_1} reports the corresponding relaxation performance.


\subsection{Robot Arm}
We consider optimizing trajectories and time scaling of a 7D double integrator operating in an environment containing a bookshelf and two bins Fig.~\ref{fig:iiwa_banner}, with LBR iiwa robot arm geometry. We consider this a step towards reasoning about the full manipulator dynamics, and plan to address the remaining challenges in future work.

\begin{figure}
    \centering
\includegraphics[width=0.45\textwidth]{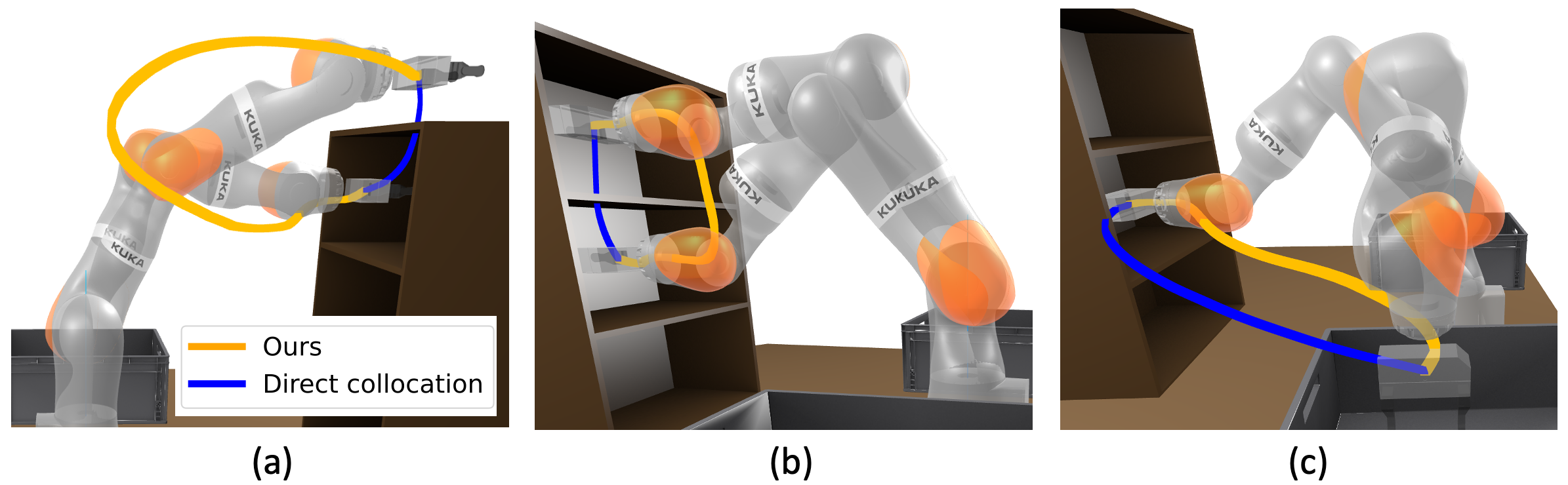}
        \caption{Minimum-time trajectories for the robot arm. The trajectories generated by our method (orange) successfully navigate the constrained environment, while those computed using direct collocation (blue) constantly result in collisions.Videos of the trajectories can be found at \href{https://lujieyang.github.io/ctfoc/comparison_AS_TS.html}{(a)}, \href{https://lujieyang.github.io/ctfoc/comparison_TS_CS.html}{(b)}, \href{https://lujieyang.github.io/ctfoc/comparison_CS_LB.html}{(c)}.}
        \label{fig:iiwa_traj_3}
\vspace{-0.2cm}
\end{figure}

The obstacle-free configuration space for the 7 DoF LBR iiwa arm is approximately decomposed into 10 convex regions using the IRIS-NP algorithm~\cite{petersen2023growing}.
Our method determines the optimal sequence of convex regions to traverse, as well as the optimal trajectories within each region.
We compare TFR with direct collocation~\cite{kelly2017introduction} on 5 tasks for moving from different initial to final configurations. While direct collocation imposes obstacle avoidance as nonlinear constraints, our approach inherently handles these constraints by operating within the convex regions.
Fig.~\ref{fig:iiwa_traj_3} visualizes the robot arm's time-optimal trajectories for moving (a) from above the shelf to the top shelf, (b) from the top shelf to the middle shelf, (c) from the middle shelf to the left bin.
In all tasks, direct collocation fails to find collision-free paths, even when initialized with the trajectories produced by TFR. Previous GCS and sampling-based methods were also unable to find optimal trajectories that strictly adhere to input limits with a time penalty. Our proposed formulation consistently produces empirically tight solutions as reported in Tab.~\ref{tab:iiwa_opt_gap}, leading to fast, smooth, and collision-free trajectories that respect the dynamical constraints in complex environments. 
% \russt{there aren't really any (quantitative) results presented here? i think we need to state more clearly whether/why we consider this example a success. just the fact that it runs at all? it can do something that other gcs trajopt transcriptions can't (you should say that). does it do them well?}
% \begin{figure}
% \centering
% \includegraphics[width=0.22\textwidth]{figures/iiwa_opt_gap.png}
%     \caption{Relaxation gap of the TFR for moving the robot arm to different configurations in minimum time.}
% \label{fig:iiwa_opt_gap}
% \end{figure}
\begin{table}[t]
\centering
\begin{tabular}{ |c|c|c|c|c|c| } 
 \hline
 Task & 0 & 1 & 2 & 3 & 4 \\ 
  \hline
 Relaxation gap $\delta$ (\%) & 0.9 & 2.1 & 14.3 & 24.4 & 5.8 \\ 
 \hline
\end{tabular}
\caption{Relaxation gap of the TFR for moving the robot arm to different configurations in minimum time.}
\label{tab:iiwa_opt_gap}
\vspace{-0.2cm}
\end{table}
\section{Conclusion}
In this paper, we propose a novel SDR for the optimal control of linear and PWA systems with time scaling. This framework utilizes techniques from semidefinite programming, exploits the structure of the OCP, and enables the efficient computation of globally optimal solutions.
Our formulation allows optimizing over the time scaling and the state and control trajectories jointly, in contrast to common bilevel methods.
In addition, our TFR can be integrated into the GCS framework for mode optimization. This enables the joint optimization of the mode sequence and continuous system trajectories within a unified convex program. 

\appendix
\label{sec:appendix}

This appendix details the proposed semidefinite program with the additional constraints that tighten our TFR.
We consider the following abstract representation for the nonconvex OCP with time scaling for linear systems \eqref{eq:lti_dynamics}:
\begin{subequations}\label{eq:lti_abstract}
\begin{align}
    \minimize \ & \eta Kh + \tr((Q_x + R_u) \cdot hrr^\top)\\
    \text{subject to} \ & Fr = Ghr \quad\;\; \text{(dynamical constraints)}\\
    & Cr + d \ge 0  \quad\; \text{(input and state constraints)} \\
    & Hr = b  \quad\quad\quad \text{(terminal conditions)}\\
    & h \geq 0.
\end{align}
\end{subequations}
Our SDR reads
\begin{subequations}\label{eq:lti_tsr}
\begin{align}
    \minimize \ & \eta K h  + \tr((Q_x + R_u) \cdot hrr^\top) \\
    \text{subject to} \
    & Fr = Ghr, \
    F r/ h = Gr\\
    & Fr r^\top = Gh r r^\top, \
    F rr\top/h = G r r^\top \\
    \label{eq:12f}
    & Hr = b , \
    H h r = bh  \\
    & H r/ h = b / h,  \
    Hrr^\top = br^\top  \\
    & Hhrr^\top = bhr^\top, \
    H rr\top/ h = b r^\top / h  \label{eq:12k}\\
    & C r + d \ge 0, \
    (C r + d)(C r + d)^{\top}\ge 0 \\
    & h(C r + d) \ge 0, \
    h(C r + d)(C r + d)^{\top} \ge 0 \\
    & (C r + d) / h \ge 0, \
    (C r + d)(C r + d)^{\top} / h \ge 0 \\
    &h\ge 0, \ 1/h \geq 0,
\end{align}
\end{subequations}
where the products of variables are replaced by the corresponding entries from the matrix $Y$ in~\eqref{eq:change_of_variable}.
The additional equality constraints necessary for the sparse TFR are 
\begin{equation}\label{eq:state_consistency_additional}
\begin{aligned}
    hr_k\left[n_x : 2 n_x\right] &= hr_{k+1}[:n_x] \\
    \frac{r_k}{h}\left[n_x : 2 n_x\right] &= \frac{r_{k+1}}{h}[:n_x] \\
    r_kr_k^\top\left[n_x : 2 n_x, n_x : 2 n_x\right] &= r_{k+1}r_{k+1}^\top[:n_x, :n_x] \\
    hr_kr_k^\top\left[n_x : 2 n_x, n_x : 2 n_x\right] &= hr_{k+1}r_{k+1}^\top[:n_x, :n_x]\\
    \frac{r_kr_k^\top}{h}\left[n_x : 2 n_x, n_x : 2 n_x\right] &= \frac{r_{k+1}r_{k+1}^\top}{h}[:n_x, :n_x].
\end{aligned}
\end{equation}

\bibliographystyle{IEEEtran}
\bibliography{reference.bib}

\end{document}